\documentclass[sigconf, authorversion]{acmart}

\copyrightyear{2026}
\acmYear{2026}
\setcopyright{cc}
\setcctype{by}
\acmConference[ICAIF '26]{7th ACM International Conference on AI in Finance}{November 14--17, 2026}{Milan, Italy}
\acmBooktitle{7th ACM International Conference on AI in Finance (ICAIF '26), November 14--17, 2026, Milan, Italy}
\acmDOI{10.1145/3858063.3858904}
\acmISBN{979-8-4007-3087-0/2026/11}

\begin{document}

\title{Internalizing Investment Mandates: \\ A Neurosymbolic RL Approach to Portfolio Allocation}

\author{Zeynep Y{\i}ld{\i}z}
\email{zeynep.yildiz@std.bogazici.edu.tr}
\orcid{0009-0006-4638-2632}
\affiliation{%
  \institution{Bo\u{g}azi\c{c}i University}
  \city{Istanbul}
  \country{T\"{u}rkiye}}
\affiliation{%
  \institution{Finesse Lab, University of Amsterdam}
  \city{Amsterdam}
  \country{the Netherlands}}

\author{Satchit Chatterji}
\orcid{0009-0003-8648-1158}
\email{s.chatterji@uva.nl}
\affiliation{%
  \institution{Finesse Lab, University of Amsterdam}
  \city{Amsterdam}
  \country{the Netherlands}}

\author{Ercan Atam}
\orcid{0000-0002-6073-5143}
\email{ercan.atam@bogazici.edu.tr}
\affiliation{%
  \institution{Bo\u{g}azi\c{c}i University}
  \city{Istanbul}
  \country{T\"{u}rkiye}}

\author{Erman Acar}
\orcid{0009-0002-9656-7249}
\email{erman.acar@uva.nl}
\affiliation{%
  \institution{Finesse Lab, University of Amsterdam}
  \city{Amsterdam}
  \country{the Netherlands}}

\begin{abstract}
Institutional portfolios operate under investment policy statements that specify asset allocation constraints, often conditioned on market regime information. Deep reinforcement learning has been shown to optimize financial performance but typically does not enforce rule compliance, making it unsuitable for real-world deployment. Existing neurosymbolic methods such as Probabilistic Logic Policy Gradient (PLPG) offer constraint-satisfying reinforcement learning (RL) via differentiable symbolic supervision but assume a finite, discrete action space. We present a PLPG-inspired continuous allocation method that estimates the probability that the policy's own executed action satisfies the investment mandate, using reparameterized Monte Carlo sampling, and adds it to the training objective. No post-hoc enforcement mechanism intercepts actions at inference time and compliance is learned by the agent. Audited directly on deterministic mean actions with no correction, the policy reduces violations of a six-rule regime-conditional mandate from 96-98\% for unconstrained RL and 33.4\% for a reward-penalty baseline to 4.8\%, showing that a continuous-action neurosymbolic policy can substantially internalize a symbolic mandate.
\end{abstract}

\ccsdesc[500]{Computing methodologies~Reinforcement learning}

\keywords{safe reinforcement learning, portfolio optimization, neurosymbolic AI,
market regime detection, constrained MDP}

\maketitle

\section{Introduction}
\label{sec:intro}


\begin{figure*}[t]
  \centering
  \includegraphics[width=\textwidth]{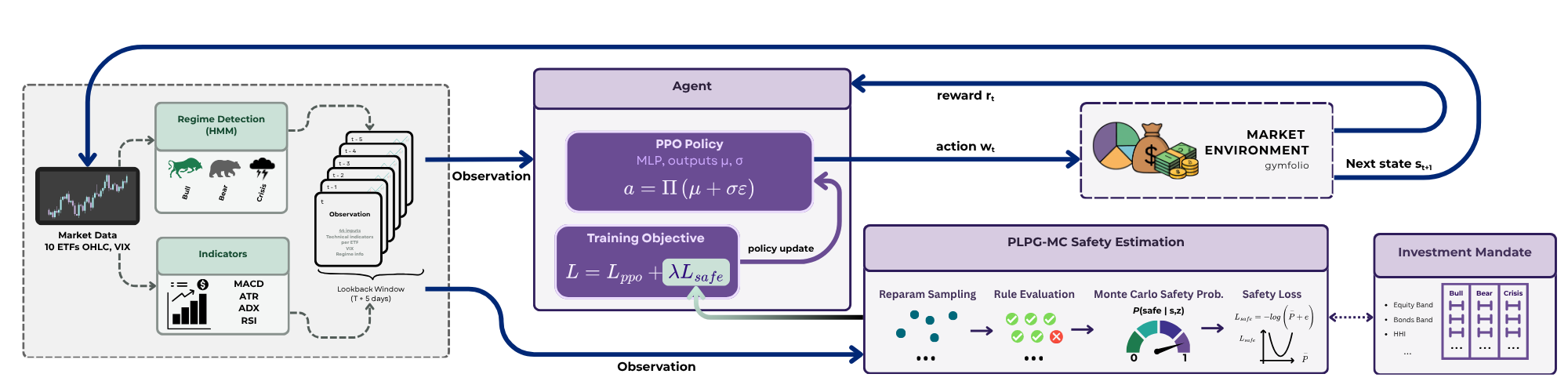}
  \caption{Overview of the proposed framework.}
  \Description{Diagram of the framework: an HMM regime detector and market features feed a PPO policy; a Monte Carlo safety estimator scores sampled allocations against the regime's mandate and adds a safety loss to the PPO update.}
  \label{fig:arch}
\end{figure*}

Institutional portfolios are managed under an \textit{investment policy statement}: a document agreed in advance between the portfolio manager and the client that specifies the eligible asset classes and an allocation band around each of them ~\cite{maginn2007managing}. These requirements are naturally expressed as symbolic rules, written in human readable form and verifiable directly from the portfolio holdings alone. A compliance function independently verifies orders before execution and holdings afterward. A mandate breach is considered a compliance event regardless of quarterly results and cannot be offset by good financial performance. Those bands, however, are rarely fixed. Market conditions can alter the dependence structure and risk characteristics of assets ~\cite{angbekaert2002regime}, motivating investment guidelines whose applicable constraints vary with the current market regime.

Trained purely to maximize risk-adjusted returns, reinforcement-learning agents have no inherent mechanism to respect such mandates. Consequently, unconstrained learners are known to breach portfolio risk limits routinely~\cite{enkhsaikhan2025augmented}. Constrained MDP methods are the standard remedy which bound the expected cost of a policy, implicitly creating an exchange rate that permits violations to be paid for with high returns. A real-world mandate admits no such exchange rate. What matters is whether the executed allocation satisfies the rules on any given day (a hard, \emph{yes} or \emph{no} per-day), so compliance is a property that must be measured directly.

This setting is well-suited for neurosymbolic AI, combining symbolic domain knowledge with neural learning so that symbolic rules can directly shape the behaviour of the learned model~\cite{kautz2022thirdsummer}. Its use in finance, however, is limited~\cite{bouneffouf2022neurosymbolic} and even rarer in RL, where symbolic structures typically serve as input features instead of enforced constraints~\cite{harini2023neurosymbolic}. To our knowledge, no prior work internalizes a regime-conditional mandate in a continuous-action portfolio policy.

In safe RL, a common approach is \textit{shielding}, which consists of placing a \emph{safety layer} between the agent's policy and the environment such that unsafe actions are corrected before they are executed ~\cite{alshiekh2018shielding,chatterji2025analyzing}. Probabilistic logic shielding uses probabilistic semantics in order to interpret the specification set as differentiable, enabling constraint satisfaction to be directly optimized in the loss function using common gradient-based methods such as gradient descent.
This approach is known as \emph{Probabilistic Logic Policy Gradients} (PLPG)~\cite{yang2023plpg}. It is built for discrete action spaces, where the shield adjusts each of the agent's actions by their estimated safety. This function has a closed form state-safety normalizer (the sum of action-conditioned safety over each action). However, portfolio allocation is in general continuous, thus that normalizer becomes an intractable integral over the policy simplex. 
One route for continuous control recovers tractability by learning an agent–environment model, which is costly and can introduce errors ~\cite{goodall2023ambs,goodall2024continuous}.

The type of portfolio agent studied here is a \textit{price-taker}: the executed allocation is exactly the weight vector the agent chooses. More importantly, the mandate is a predicate on that vector and not on any resulting market state. Each rule defines a simple geometric constraint on those weights e.g., comparing a sum of weights or their squared norm to a threshold and can be evaluated in closed form. The safety quantity depends only on the agent's action; what remains is the expectation over the policy, which we estimate by sampling.
The resulting term, PLPG-MC, is a reparameterized Monte Carlo surrogate for the policy's own safety probability, added to PPO. A hidden Markov model with three states supplies the market regime that selects which thresholds are active.

This paper makes the following contributions towards RL-based portfolio allocation under investment mandates: 
\begin{itemize}
  \item We develop a PLPG-inspired continuous-action safety objective that replaces the discrete normalizer with a reparameterized Monte Carlo estimate of the base policy’s executed-action safety.
  \item We perform extensive experiments using a regime-conditioned mandate over portfolio weights, audited on the policy alone, and present a careful ablation study on the effect of differentiable rules and reward penalties.
  \item We study the effect of the size of overlapping safe regions of different rules defined in the mandate set.
\end{itemize}

\section{Preliminaries and Problem Formulation}
\label{sec:background}

This section defines the allocation environment, mandate, and violation measurement under the regime-conditional constrained MDP.

\subsection{Constrained MDPs}
A \emph{Markov Decision Process} (MDP) is a tuple $\langle\mathcal{S}, \mathcal{A}, P, R, \gamma\rangle$ consisting of states $s\in \mathcal{S}$, actions $a \in \mathcal{A}$, transition kernel $P(s' \mid s, a)$ over next states, reward function $R:\mathcal{S}\times\mathcal{A}\rightarrow\mathbb{R}$ and discount factor $\gamma\in[0,1)$. We consider a stochastic policy $\pi_\theta(a \mid s)$ parameterized by $\theta$ trained to maximize expected discounted return over an infinite horizon, i.e.\ $J(\theta)=\mathbb{E}_{\pi_\theta}\left[\sum_{t=0}^\infty \gamma^t R_t\right]$~\cite{sutton2018rl}. \emph{Policy gradient} (PG) methods maximize $J(\theta)$ by estimating $\nabla_\theta J(\theta)$ from sampled trajectories and ascending it. We use \emph{Proximal Policy Optimization} (PPO)~\cite{schulman2017ppo}, which takes several gradient optimization steps on each batch of collected experience and clips the ratio between the updated policy and the one that collected the batch, so no single update moves the policy far from the data it was previously fit on. \emph{Constrained MDPs} (CMDPs) augment the MDP setting with a cost signal $C:\mathcal{S}\times\mathcal{A}\rightarrow\mathbb{R}$ and a budget $d\in \mathbb{R}$, and common treatments, such as Lagrangian relaxation and constrained policy search, aim to bound the \emph{expected cost} of the policy under $d$~\cite{altman1999cmdp,achiam2017cpo}.

\subsection{The Portfolio Allocation MDP} \label{sec:portfolio-mdp}
\paragraph{Environment}  We build on \emph{gymfolio} ~\cite{espiga2025gymfolio}, a portfolio optimization environment whose \texttt{PortfolioOptimizationEnv} separates Open-High-Low-Close (OHLC) price data from a second stream of per-day observations (technical indicators, the VIX and, in our case, the annotated market regime \S\ref{sec:hmm}). It additionally rebalances on a fixed schedule and exposes the observation windows as a stacked frame. We rebalance daily and configure the reward as the realized daily portfolio return $R_t$.
\paragraph{Actions} In general, the agent allocates across $|\mathcal{A}|$ assets, and so an action $a \in \mathcal{A}$ is a continuous vector, where each element corresponds indirectly to a preliminary order for a single asset. In our case, the agent is allowed $|\mathcal{A}|=10$ ETFs: five equities (SPY, QQQ, IWM, XLF, XLE), three fixed income assets (TLT, IEF, LQD), and two commodities (GLD, SLV). The policy is a diagonal Gaussian on the raw action $a \in \mathbb{R}^{10}$, $\pi_\theta(a \mid s) = \mathcal{N}\!(\mu_\theta(s),\, \operatorname{diag}(\sigma_\theta^{2}))$, where a neural network outputs the mean $\mu_\theta(s)$ and $\sigma_\theta$ is a learned parameter shared across all states, so $a = \mu_\theta(s) + \sigma_\theta \odot \varepsilon$ with $\varepsilon \sim \mathcal{N}(0,I)$ for exploration and $\odot$ the element-wise product. The environment executes $w = \Pi(a)$, where $\Pi$ clamps each coordinate to $[0,1]$ and divides by the sum, thus $w$ is a valid distribution (specifically, $w\in \Delta^{9}$). The book is fully invested with no cash, leverage or short positions, and the mandate is evaluated on $w$.

\paragraph{States} The observed state includes 44 features: 41 market features (four technical indicators per asset and the VIX level) together with the one-hot encoding of the estimated market regime $z$ (\S\ref{sec:hmm}). The policy sees a six day frame (the current day and five prior) so the input is $6 \times 44 = 264$ values. Regime-blind ablation uses the 41 market features alone, $6 \times 41$ values. The technical indicators are as follows: MACD (12/26/9 signal line), RSI (30-day), ADX (30-day) and ATR (14-day). The policy receives the decoded regime label and not the detector's posterior, so it observes no uncertainty at transitions (\S\ref{sec:discussion}). 

\begin{table}[t]
  \centering
  \caption{The regime-conditional mandate: thresholds on the executed
  weight vector by regime.}
  \label{tab:mandate}
  \begin{tabular}{lccc}
    \toprule
    Rule & Bull & Bear & Crisis \\
    \midrule
    Max HHI (concentration) & 0.35 & 0.28 & 0.40 \\
    Min equity              & 0.40 & 0.15 & 0.00 \\
    Max equity              & 0.80 & 0.50 & 0.30 \\
    Min bonds               & 0.10 & 0.25 & 0.30 \\
    Max bonds               & 0.40 & 0.60 & 1.00 \\
    Min commodity           & 0.00 & 0.05 & 0.15 \\
    \bottomrule
  \end{tabular}
\end{table}

\paragraph{Constraints} Our constraints are selected by an exogenous \emph{market regime} $z(s) \in \{\text{Bull}, \text{Bear}, \text{Crisis}\}$: a persistent, latent state in which returns, volatilities and cross-asset correlations share a common character and which gives way to another at unpredictable intervals \cite{hamilton1989regime, angbekaert2002regime}. We recover it with a hidden Markov model (\S\ref{sec:hmm}) and pass the decoded label into the observation, so $z$ is a deterministic function of $s_t$. We write $z$ for $z(s)$ throughout for brevity. 

Let $\mathcal{R}$ be the set of strict (Boolean) natural-language rules listed by some institutional mandate. From these, logical constraints are constructed, denoted $c_r(w,z)$ ($\forall r\in\mathcal{R}$), whose satisfaction is based on $z$ and the executed portfolio weights $w$.
 Each constraint $c_r$ compares a measurement $g_r(w)$ against a predefined threshold $\tau_r(z)$ that depends on the regime, either as a ceiling $g_r(w) \le \tau_r(z)$ or a floor $g_r(w) \ge \tau_r(z)$. 
The \emph{safe set} $\mathcal{W}(z)$ (i.e.\ mandate-following allocation set) is the intersection over $\mathcal{R}$ of the portfolio weights $w$ each rule admits:
\begin{equation}
\label{eq:safeset}
\mathcal{W}(z) \;=\; \bigcap_{r \in \mathcal{R}} \bigl\{\, w \in \Delta^{9} : c_r(w, z) \,\bigr\}.
\end{equation}

Our object of study is a policy whose own executed actions lie in $\mathcal{W}(z_t)$ at deployment, with no additional safety machinery. Traditional CMDP expected cost formulations would generally permit some level of non-zero trading violations in this setting against return, but a mandate does not, which is why we treat the rules as hard constraints and count a day as violating the mandate whenever $w \notin \mathcal{W}(z)$.

\paragraph{Mandate.} Table~\ref{tab:mandate} lists the six rules we consider here. Herfindahl-Hirschman Index (HHI) is the concentration score $\sum_i w_i^2$; the remaining rules are floors (or `Min.') and caps (or `Max.') on the equity, bond and commodity sleeves. A floor or cap on a single measurement $g_r(w)$ is a regime-conditional predicate for each entry in Table~\ref{tab:mandate}. The Max-equity rule in Bull reads \emph{`if Bull, then $\sum_{i \in \mathrm{eq}} w_i \le 0.80$'}, and the bond floor in Crisis reads \emph{`if Crisis, then $\sum_{i \in \mathrm{bd}} w_i \ge 0.30$'}; HHI caps the concentration score $\sum_i w_i^2$ the same way. Equity capacity contracts from Bull to Crisis ($0.80 \to 0.50 \to 0.30$) while the bond floor ($0.10 \to 0.25 \to 0.30$) and commodity floor ($0 \to 0.05 \to 0.15$) rise. Two design choices increase flexibility under stress: the Crisis HHI cap is looser than Bear ($0.40$ vs.\ $0.28$), and the commodity sleeve has a floor but no cap, allowing greater concentration in defensive assets when correlations rise. The thresholds form a representative stylized mandate with sleeve-level bands and concentration limits similar to those used in regulated funds~\cite{ucits2009}.

\paragraph{Constraint geometry.} By construction, no static allocation satisfies all three regimes at once. Bull requires at least 40\% equity and Crisis permits at most 30\%, so $\mathcal{W}(\mathrm{Bull}) \cap \mathcal{W}(\mathrm{Bear}) \cap \mathcal{W}(\mathrm{Crisis}) = \emptyset$ and thus, compliance requires reallocation. 

\paragraph{Evaluation metrics.} All results are measured on the \emph{held out} test period: the model is fit on data from 2005 to 2020 and evaluated on 2021-2026. Deployment evaluation is a \emph{passive audit}: the trained policy is rolled through this period acting deterministically at its mean, $w_t = \Pi(\mu_\theta(s_t))$, with $\Pi$ as defined in \S\ref{sec:portfolio-mdp} and with no active correction to the action. We report percent violation (PV) as the percentage of $T$ test days on which the executed action violates the active ruleset:
\begin{equation}
\label{eq:passive}
\mathrm{PV} \;=\; \frac{100}{T} \sum_{t=1}^{T}
\mathbf{1}\!\left[\, w_t \notin \mathcal{W}(z_t) \,\right],
\end{equation}
We also report the total return over the test period, Sharpe ratio, maximum drawdown and daily $\mathrm{CVaR}_5$. Sharpe uses a zero risk-free rate, applied identically to every strategy.

\begin{figure*}[t]
  \centering
  \includegraphics[width=0.85\textwidth]{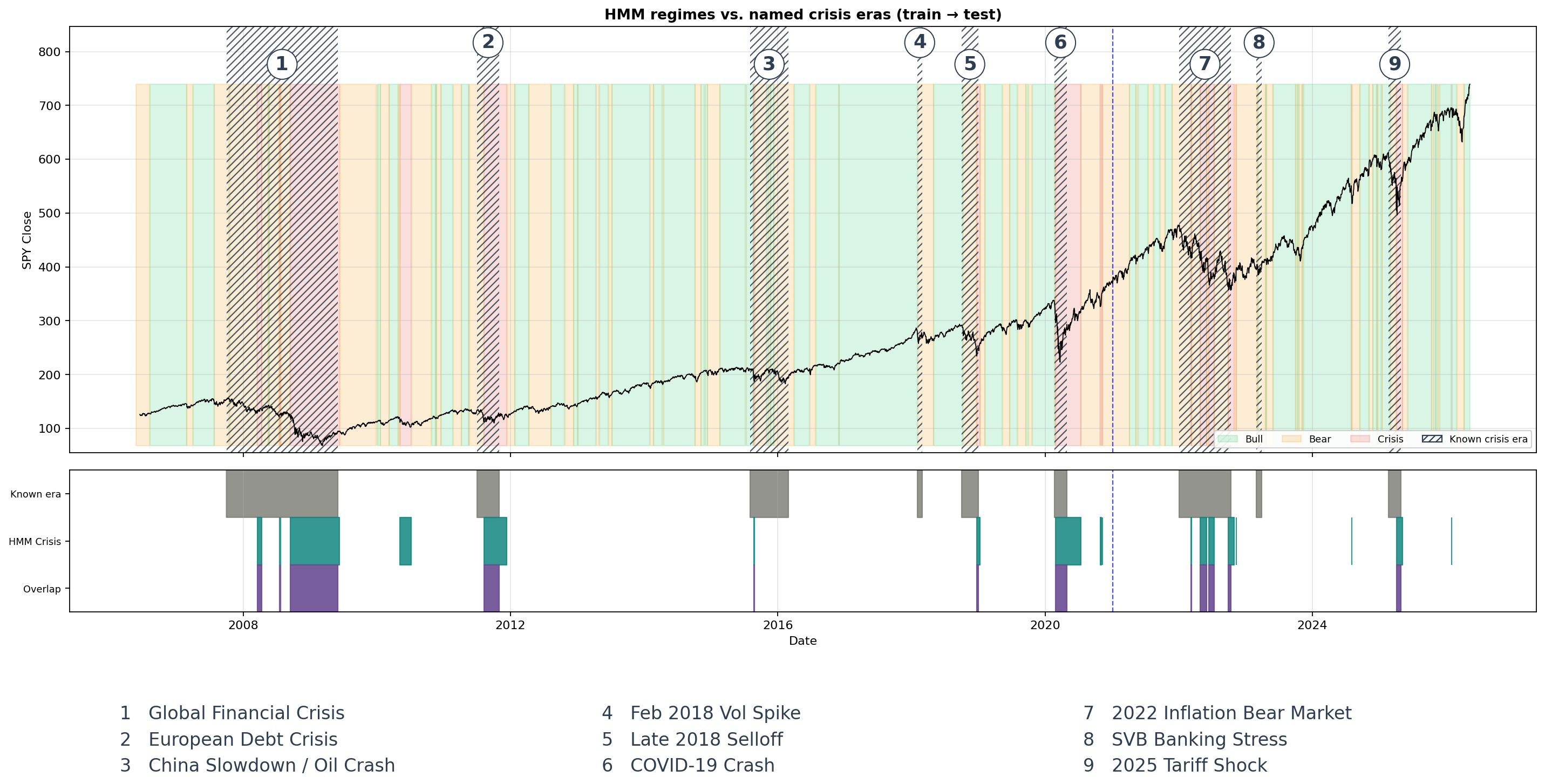}
  \caption{HMM regime detection results against specified stress periods.}
  \Description{SPY price over time, shaded by the HMM's Bull, Bear and Crisis regimes, with hatched bands marking nine named crisis periods.}
  \label{fig:market_regime}
\end{figure*}
 
\section{Method}
\label{sec:method}

Our method adds one term to the PPO training objective: a differentiable estimate of the probability that the policy's own executed actions satisfy the enforced mandate. We call this method \textit{PLPG-Monte Carlo} (PLPG-MC). Everything else is held fixed so the method is separable from its baselines in ablation of \S\ref{sec:experiments}.

\subsection{HMM Market-Regime Detection}
\label{sec:hmm}
A \emph{hidden Markov model} (HMM) ~\cite{hamilton1989regime}, a latent discrete state evolving as a Markov chain with conditionally generated observations, supplies the regime prediction $z \in \{\text{Bull}, \text{Bear}, \text{Crisis}\}$. It is fit on eight causal stress features spanning the universe (equity return, cross-asset dispersion, realized volatility, drawdown, a credit proxy, commodity return, VIX, and rolling equity bond correlation), each z-scored with training statistics only and decoded with forward filtered posteriors, so the label at time $t$ uses only information available at $t$. A two-component Gaussian mixture emission is selected over full and diagonal Gaussians on separation in stress space, crisis share and Bayesian Information Criterion (BIC). Since HMM states have no intrinsic labels, we assign them from per state statistics, Crisis being the regime of highest mean VIX, and Bull the regime of higher mean equity return among the remaining two. During the training period, the three regimes exhibit a clean stress ordering (annualized equity volatility 9.4\%, 18.1\%, and 44.4\%). Against the Office of Financial Research (OFR) Financial Stress Index (FSI), built daily from 33 market variables by the U.S. Office of Financial Research~\cite{ofr2026fsi}, the three states order cleanly over 5018 days (medians $-2.38$, $-0.11$ and $+4.89$, Spearman $\rho = 0.65$), and the ordering also holds on that index's funding subindex, which our features do not observe (Figure~\ref{fig:market_regime}). The regime enters the method in two places: as a component of the policy observation and as the selector of the active threshold set $\tau_r(z)$ in \eqref{eq:slack}.

\subsection{Neurosymbolic Regularization}
For each rule $r$ and regime $z$, we define the signed slack:
\begin{equation}
\label{eq:slack}
\mathrm{slack}_r(w, z) =
\begin{cases}
\tau_r(z) - g_r(w) & \text{(for ceiling rules)} \\
g_r(w) - \tau_r(z) & \text{(for floor rules)}
\end{cases}
\end{equation}
where $\tau_r(z)$ is the regime-conditional threshold and $g_r$ the measurement function (e.g., $g_{\mathrm{HHI}}(w) = \sum_i w_i^2$, $g_{\mathrm{eq}}(w) = \sum_{i \in \mathrm{eq}} w_i$). To keep the safety term from pulling the policy deeper into the feasible set than the mandate requires we define the \emph{effective} slack by clipping the slack on its safe side at a margin $m$:
\begin{equation}
\label{eq:margin}
\mathrm{slack}_r^{\mathrm{eff}}(w,z) = \min\!\left(\mathrm{slack}_r(w,z),\, m\right), \qquad m = 0.02,
\end{equation}
So a rule stops contributing gradient once it is satisfied by $m$, a dead-zone in the safe interior. Without it, the sigmoid below supplies gradient arbitrarily far inside the feasible set. A temperature-$\beta$ logistic $\mathrm{sigm}$ relaxes the indicator into a per-rule safety $P_r \propto \mathrm{sigm}(\beta\,\mathrm{slack}_r^{\mathrm{eff}})$, normalized so a rule satisfied by the margin scores exactly $1$. The joint safety is the product of these scores (the soft relaxation of the conjunction $\Pi_r1[c_r(w,z)]$), whose logarithm is
\begin{equation}
\label{eq:logsafe}
\log P(\mathrm{safe} \mid w, z) = \textstyle\sum_{r}\bigl[\mathrm{softplus}(-\beta m) - \mathrm{softplus}(-\beta\,\mathrm{slack}_r^{\mathrm{eff}})\bigr].
\end{equation}

\subsection{Continuous Actions: A Reparameterized Monte Carlo Surrogate}
\label{sec:estimator}
The construction of Yang et al.~\cite{yang2023plpg} trains through a shielded policy $\pi^{+}$ whose normalizer is a finite sum over actions in their discrete setting. Over the portfolio simplex $\Delta^{9}$, the integral has no closed form, so $\pi^{+}$ cannot be constructed and the method does not transfer as stated. We target the same quantity on the base policy which is the probability that its own executed action is safe. Through the reparameterization of \S\ref{sec:background}, this is 
\begin{equation*}
P_{\pi_\theta}(\mathrm{safe}\mid s) = \mathbb{E}_{\varepsilon\sim\mathcal{N}(0,I)}\!\big[P(\mathrm{safe}\mid \Pi(\mu_\theta(s)+\sigma_\theta\odot\varepsilon),\, z)\big],
\end{equation*}
the executed-action safety that the deployment audit of Eq.~\eqref{eq:passive} measures because $\pi_\theta$ is what acts.
Pursuing it in a continuous action space poses two separate difficulties: the per-action safety $P(\mathrm{safe} \mid w, z)$ must be computable and differentiable and the expectation over the policy must be tractable.

A world model~\cite{goodall2023ambs,goodall2024continuous}, a grid over $\Delta^9$ and runtime correction each buy tractability at a cost we can avoid. The price taking assumption of \S\ref{sec:background} dissolves the first difficulty (computing and differentiating $P(\mathrm{safe}\mid w,z)$) without any of them because the agent does not move prices. Equation~\eqref{eq:logsafe} evaluates safety exactly from the executed weights, leaving only the policy expectation to be estimated by sampling.

At state $s$ with regime $z$ we draw $k\in\{1,...,K\}$ reparameterized action samples
\begin{equation}
\label{eq:samples}
w_k = \Pi\!\bigl(\mu_\theta(s) + \sigma_\theta \odot \varepsilon_k\bigr),
\qquad \varepsilon_k \sim \mathcal{N}(0, I),
\end{equation}
where $\Pi$ is the clamp-and-normalize projection of \S\ref{sec:background}, so safety is evaluated on exactly the actions the environment would execute. The clamp has zero slope outside $[0,1]$, which makes single-asset corners absorbing under any $\lambda$, so we pass a straight-through gradient through it and keep the exact gradient of the normalization. We then define the estimated safety as:
\begin{equation}
\label{eq:loss}
\widehat{P}(s) = \tfrac{1}{K} \textstyle\sum_{k=1}^{K} P(\mathrm{safe} \mid w_k, z)
\end{equation}
We use $K = 8$. 
$\widehat{P}$ is an unbiased estimate of the executed-action safety $P_{\pi_\theta}(\mathrm{safe} \mid s)$. Via a semantic loss \cite{xu2018semanticloss}, we transform this into a differentiable regularizer (`safety loss') as:
\begin{equation}
\label{eq:safety_loss}
    \mathcal{L}_{\mathrm{safe}}(s) = -\log \widehat{P}(s)
\end{equation}
The gradient of $\mathcal{L}_{\mathrm{safe}}(s)$ reaches $\mu_\theta$ through the reparameterization path \cite{kingma2014auto}, at lower variance than a likelihood-ratio estimate. It would also reach $\sigma_\theta$, but we block that path with a stop-gradient (\emph{sigma-detach}, below) so the safety term acts on the mean alone. Taking the logarithm introduces bias. By Jensen's inequality $\mathbb{E}[\log \widehat{P}] \le \log \mathbb{E}[\widehat{P}] = \log P_{\pi_\theta}(\mathrm{safe}\mid s)$, so $\mathbb{E}[\mathcal{L}_{\mathrm{safe}}]$ upper-bounds the true negative log-safety and training minimizes a \emph{conservative surrogate}. The bias shrinks with $K$ and is shared by Monte Carlo log-likelihood estimators generally. We use PLPG-MC to distinguish the Monte Carlo surrogate from discrete PLPG and make no stronger equivalence claim, since neither PLPG’s convergence guarantee nor its objective equivalence survives the loss of $\pi^{+}$
 (cf.\ \S\ref{sec:discussion}).

\paragraph{Controlling variance collapse.} The safety loss carries a gradient towards smaller policy variance. Once $\mu_\theta$ sits inside the feasible set, concentrating samples around it also raises $\widehat{P}$ so lowering $\sigma_\theta$ becomes a way to reduce violations. Since our log-std is a global parameter shared across states, this could reduce exploration everywhere. There are two design choices we made to close this pathway.
Per Eq.~\eqref{eq:margin}, the margin $m$ removes the residual pull towards the interior of the feasible set once the policy satisfies the mandate. 
\emph{Sigma-detach} closes the main route to collapse. Because $\widehat{P}$ rises as samples concentrate around a feasible mean, the safety loss can be lowered by shrinking $\sigma_\theta$ rather than by moving $\mu_\theta$; with $\sigma_\theta$ a single global parameter, that pressure would drain exploration at every state. We therefore stop the safety gradient through $\sigma_\theta$, computing $w_k = \Pi(\mu_\theta(s) + \mathrm{sg}(\sigma_\theta) \odot \varepsilon_k)$ with $\mathrm{sg}(\cdot)$ the stop-gradient operator.
The safety term can then lower $\mathcal{L}_{\mathrm{safe}}$ only by moving the mean into the feasible set, not by collapsing the variance around it, $\sigma_\theta$ is left to the entropy and return terms. 

\paragraph{Placement.} At each visited state the safety estimator draws $K$ samples from the current policy and scores their executed allocations. The execution map $\Pi$ is the same for the sampled actions and for deployment. The difference is that the deployment audit of Eq.~\eqref{eq:passive} uses the hard indicator $\mathbf{1}[w \in \mathcal{W}(z)]$, whereas training scores samples using the differentiable relaxation of Eq.~\eqref{eq:logsafe}. During deployment, we evaluate the policy deterministically using its mean action, $w_t=\Pi(\mu_\theta(s_t))$, with no correction, resampling, or reweighting layer between $\pi_\theta$ and the environment. Thus, the reported passive violation rate measures the compliance of the deterministic deployment policy, omitting post-hoc enforcement.

\subsection{Training Objective} \label{sec:shieldedppo}
The objective we minimize is $\mathcal{L}_{\mathrm{PPO}}$, the negative of the clipped-surrogate PPO objective of Eq.~(9) of \cite{schulman2017ppo}, plus the safety term of Eq.~\eqref{eq:safety_loss}:
\begin{equation}
\label{eq:total}
\mathcal{L}(\theta) \;=\; \mathcal{L}_{\mathrm{PPO}}(\theta)
\;+\; \lambda\, \mathcal{L}_{\mathrm{safe}}(\theta)
\end{equation} 
The weight $\lambda$ is a constant that controls how strongly the safety term is applied. The safety term is unclipped and acts on the same policy PPO updates. It never enters the reward, the value function or the advantage, so compliance is never discounted or credit-assigned across a trajectory.


\section{Experiments}
\label{sec:experiments}

We evaluate PLPG-MC against its baselines on the regime-conditional mandate. This section starts with describing the data, environment, the training setup. We then report the main compliance and return results across five seeds, sweep the safety weight $\lambda$, localize where the residual violations occur and examine how the mandate's geometry can shape compliance and regime responsiveness. 

\subsection{Setup}
Daily data span 2005-06-01 to 2026-05-11, with the model fit on data through 2020-12-31 and every reported result measured on the held out remainder (2021-01 through 2026-05), a test window that contains the 2022 drawdown, the 2023 banking stress and the 2025 tariff shock. Episodes are 63 trading days with daily rebalancing and the six-day observation frame of \S\ref{sec:background}, a third of training episodes start from regime-stratified points so that the rarer regimes are seen often enough to learn from.

Four learned variants span the mechanism space. \emph{Vanilla PPO} and \emph{Regime-Aware PPO} carry no safety term and differ only in whether the regime signal enters the observation, their performance shows what an unconstrained return-maximizer does against the mandate. \emph{Reward Penalty} subtracts a violation cost, weighted by $\lambda_{\mathrm{pen}}$, from the reward, the standard soft-constraint treatment. \emph{PLPG-MC} adds the differentiable safety loss of \S\ref{sec:estimator} alone. 

All variants share the observation space, network architecture, PPO hyperparameters and training budget of $1.5 \times 10^{6}$ steps. They differ only in the safety term applied during training. PPO uses an MLP policy with learning rate $10^{-4}$, 2048 steps per rollout, batch size 256, ten epochs per update, $\gamma = 0.99$, $\lambda_{\mathrm{GAE}} = 0.95$, clip range $0.2$, and entropy coefficient $0.01$. The safety estimator of \eqref{eq:loss} draws $K = 8$ samples per state with $\beta = 50$ and margin $m = 0.02$. Each variant is trained under five seeds $\{0, 42, 67, 123, 420\}$, and every reported number is a mean $\pm$ standard deviation across them.
We include fixed-weight portfolios (SPY, equal weight, and 60/40) and a rule-following heuristic as non-learned references.

\begin{table*}[t]
  \centering
  \caption{PV is the passive violation rate of \eqref{eq:passive}, PV$_{\mathrm{crisis}}$ its restriction to the 97 Crisis days, CVaR$_5$ the mean daily loss on the worst 5\% of days. Learned rows are mean $\pm$ std over five seeds.}
  \label{tab:main}
  \begin{tabular}{lrrrrrr}
    \toprule
    & PV \% & PV$_{\mathrm{crisis}}$ \% & Return \% & Sharpe & CVaR$_5$ \% & MaxDD \% \\
    \midrule
    \multicolumn{7}{l}{\emph{Fixed-weight books}} \\
    SPY buy \& hold & 100.0 & 100.0 & 100.5 & 0.85 & 2.43 & $-25.4$ \\
    Equal weight    & 7.2   & 100.0 & 69.5  & 0.88 & 1.74 & $-20.2$ \\
    60/40           & 100.0  & 100.0 & 40.9  & 0.65 & 1.70 & $-22.0$ \\
    \midrule
    \multicolumn{7}{l}{\emph{Rule-following heuristic (runtime enforcement)}} \\
    Regime box centroid & 0.0 & 0.0 & 58.1 & 0.80 & 1.69 & $-21.1$ \\
    \midrule
    \multicolumn{7}{l}{\emph{Unconstrained learners}} \\
    Vanilla PPO      & 96.2 $\pm$ 5.0 & 100.0 $\pm$ 0.0 & 115.7 $\pm$ 53.8 & 0.76 $\pm$ 0.21 & 3.08 $\pm$ 0.39 & $-32.2 \pm 3.9$ \\
    Regime-aware PPO & 97.9 $\pm$ 3.0 & 98.1 $\pm$ 3.6   & 92.2 $\pm$ 56.3  & 0.69 $\pm$ 0.25 & 2.73 $\pm$ 0.36 & $-29.4 \pm 7.7$ \\
    \midrule
    \multicolumn{7}{l}{\emph{Constrained learners}} \\
    Reward Penalty ($\lambda_{\mathrm{pen}}{=}0.1$) & 33.4 $\pm$ 7.1 & 56.5 $\pm$ 14.8 & 58.7 $\pm$ 8.7 & 0.74 $\pm$ 0.06 & 1.82 $\pm$ 0.06 & $-22.9 \pm 3.3$ \\
    PLPG-MC ($\lambda{=}0.10$) & 6.0 $\pm$ 1.8 & 14.9 $\pm$ 14.7 & 42.1 $\pm$ 4.7 & 0.63 $\pm$ 0.06 & 1.65 $\pm$ 0.03 & $-24.8 \pm 1.3$ \\
    PLPG-MC ($\lambda{=}0.15$) & 4.8 $\pm$ 2.0 & 25.2 $\pm$ 22.2 & 39.8 $\pm$ 5.7 & 0.60 $\pm$ 0.07 & 1.67 $\pm$ 0.03 & $-25.4 \pm 1.8$ \\
    \bottomrule
  \end{tabular}
\end{table*}

\subsection{Main Results} \label{sec:results-violations}
Compliance survives the removal of the safety machinery. Rolled through the test
period with nothing between the policy and the market, PLPG-MC violates the mandate on $4.8 \pm 2.0\%$ of days. Both unconstrained learners violate it on essentially every day (96.2\% and 97.9\%), and the reward penalty, given the same observations, budget and violation signal, stops at 33.4\%. Table~\ref{tab:main} reports the comparison in full. Supplying the regime signal as an observation changes nothing on its own.

\paragraph{Return cost.} The PLPG-MC agents earn around $41\%$ total return over the test period against $92$--$116\%$ for the unconstrained learners. Annualized volatility falls from $0.19$--$0.21$ to $0.12$, maximum drawdown improves from roughly $-30\%$ to $-25\%$, and Sharpe gives up about roughly $0.1$. \S\ref{sec:discussion} takes up where that gap comes from.

\paragraph{Tail risk and dispersion.} The mandate also makes bad days less bad, with daily $\mathrm{CVaR}_5$ falling from 3.08\% and 2.73\% for the unconstrained learners to 1.82\% under the reward penalty and 1.65--1.67\% for PLPG-MC configurations. Seed dispersion collapses alongside it. Total return spans $\pm 54$ and $\pm 56$ percentage points across five seeds of an otherwise identical unconstrained configuration, against $\pm 4.7$ to $\pm 8.7$ constrained, and a deployed system is trained once and run once, so that spread is a risk in itself. 

\paragraph{Depth of compliance.} The violation rate views compliance as binary. The five sleeve rules are thresholds on portfolio weight, so their slack is quantifiable. PLPG-MC keeps a median of ten percentage points of weight between the executed book and the nearest binding sleeve threshold, against seven for the reward penalty and one to three for the unconstrained learners. On $95\%$ of compliant days that margin exceeds $2.2$ percentage points of portfolio weight, and residual breaches are shallow, with a median depth of $4.6$ points against roughly $20$ unconstrained. The concentration cap never binds for the constrained policies, which retain at least $0.10$ of headroom on every test day whereas the unconstrained policies violate the concentration limit on $61$-$87\%$ of days.

\paragraph{Reference books.} The fixed-weight rates in Table~\ref{tab:main} follow from the mandate geometry alone. The rule following heuristic is the informative reference. A deterministic allocator that projects equal weight onto the feasible box of the active regime records no violation on any test day and earns 58.1\% at Sharpe 0.80, outperforming every constrained policy we train on risk-adjusted return. It is runtime enforcement in its simplest form. At each decision, it consults the rulebook and regime label, then projects the action into the active feasible set. We use this heuristic as a reference for externally enforced compliance. It provides a comparison for the learned policies, which must satisfy the mandate through training.

\begin{figure}[t]
  \centering
  \includegraphics[width=\columnwidth]{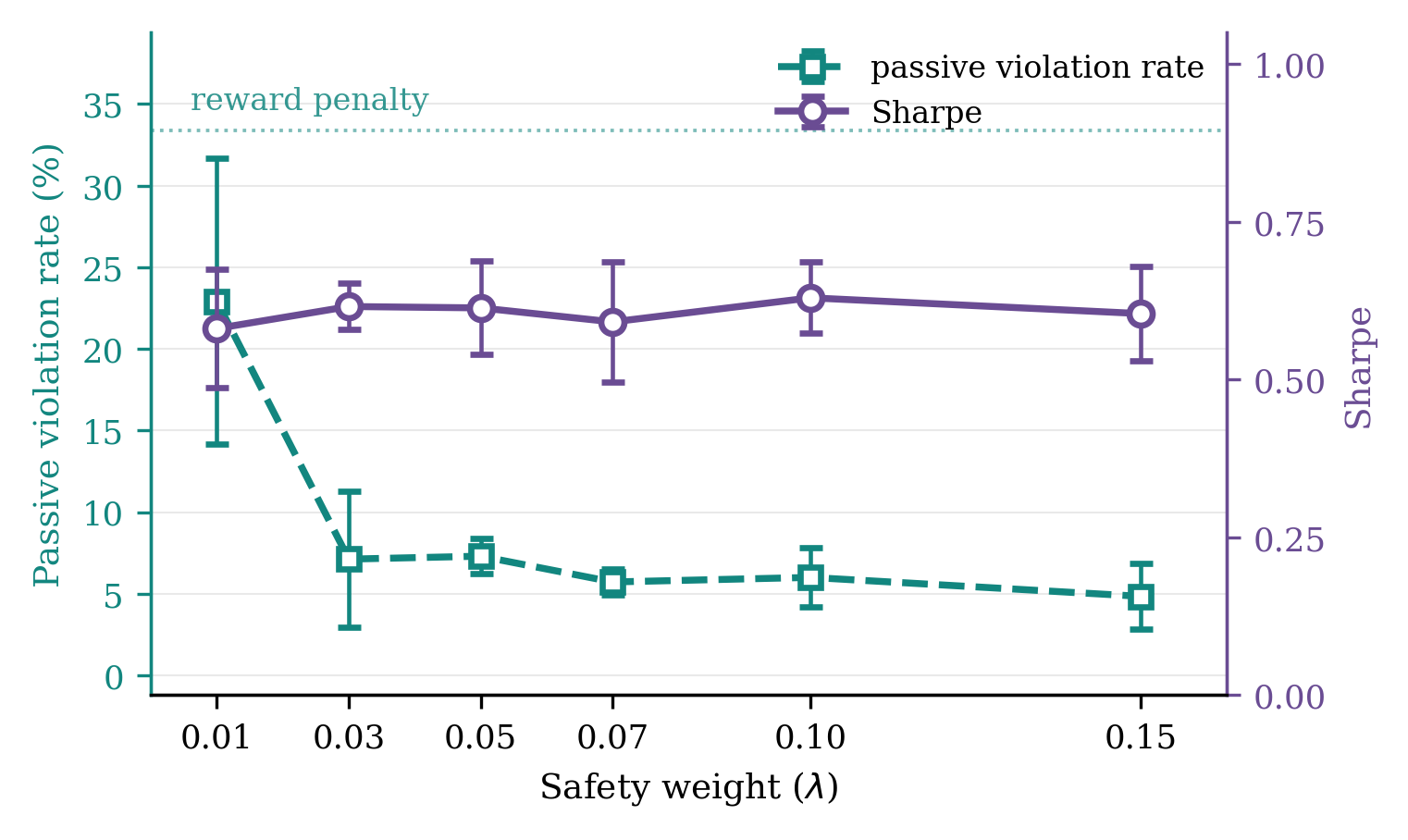}
  \caption{PLPG-MC dose--response.}
  \Description{Line plot of passive violation rate against the safety weight lambda, falling steeply up to about 0.03 and then flattening near 5 percent while Sharpe stays about the same.}
  \label{fig:dose}
\end{figure}

\subsection{Dose and Mechanism}
\label{sec:results-ablation} 

Figure~\ref{fig:dose} sweeps the safety weight $\lambda$. Violations fall steeply through $\lambda \approx 0.03$ and settle at a floor near $5\%$ that further pressure does not improve: the sweep reads $22.9$, $7.1$, $7.3$, $5.7$, $6.0$ and $4.8\%$ at $\lambda = 0.01$ through $0.15$, with the last four points inside one another's seed noise. For $\lambda \geq 0.05$ the violation rate sits on a plateau, its points indistinguishable at five seeds. We fix $\lambda = 0.15$ at $4.84 \pm 2.03$ as a representative operating point and read no within-plateau difference as meaningful.

Both mechanisms push the policy toward the feasible set, but PLPG-MC produces substantially lower residual violations under the tested configuration. Substituting it for the penalty reduces the violation rate by $28.6$ percentage points, from $33.4\%$ to $4.8\%$ (paired $t = 8.5$, $p = 0.001$, $n = 5$).

\subsection{The Residual}
\label{sec:results-regime}

Splitting violations by the regime at the decision step localizes the residual floor. Bear days are near-clean for the headline configurations (0.4--1.2\%). Bull days carry 5.7--8.8\% and Crisis days 15--25\%, with the Crisis breaches attributable to the equity cap in every variant we examined. The two failure modes are directional opposites. In Bull the binding breaches are the equity floor and the bond cap, so the policy is slow to re-risk once stress passes, and in Crisis it is slow to de-risk when stress arrives.
 
Both point at the lag of the regime filter at transitions (\S\ref{sec:hmm}), and a windowed split confirms it in part. Counting the first three days after each regime change separately, Bull entry rates run about three times the settled rate (22.3\% against 7.5\% at $\lambda = 0.10$) while in Crisis about twice (25.6\% against 12.4\%) on a settled rate of 12--22\%, so latency accounts for roughly a third of the Crisis residual and the crisis book is only partly internalized. These windows are short, 18 Crisis entry days on average, and we read them as descriptive.
 
Reallocation is not compliance on its own. The unconstrained regime policy separates its books across regimes further than any constrained variant ($L_1 = 0.55 \pm 0.14$) and moves the wrong way, raising equity from 0.55 in Bull to 0.68 in Crisis, where the constrained policy runs 0.49, 0.38 and 0.23, as prescribed by the rules.

\paragraph{Within-regime stability.}
Constrained policies hold their books more steadily inside a regime. The mean per-asset temporal standard deviation within regimes falls from $0.091$ for regime-aware PPO to $0.030$ under the reward penalty and $0.017$--$0.019$ for PLPG-MC at $\lambda \ge 0.03$. Taking the ratio of between-regime separation to within-regime dispersion as a signal-to-noise measure of mandated behavior, the unconstrained and reward-penalty policies both score $6.2$ while PLPG-MC rises with dose: $14.2$, $17.5$, $17.2$, $20.3$, $21.8$ and $22.7$ at $\lambda = 0.01$ through $0.15$. Constrained agents concentrate their behavioral variation at regime boundaries, which is the behavior a regime-conditional mandate asks for.

\begin{figure}[t]
  \centering
  \includegraphics[width=0.85\columnwidth]{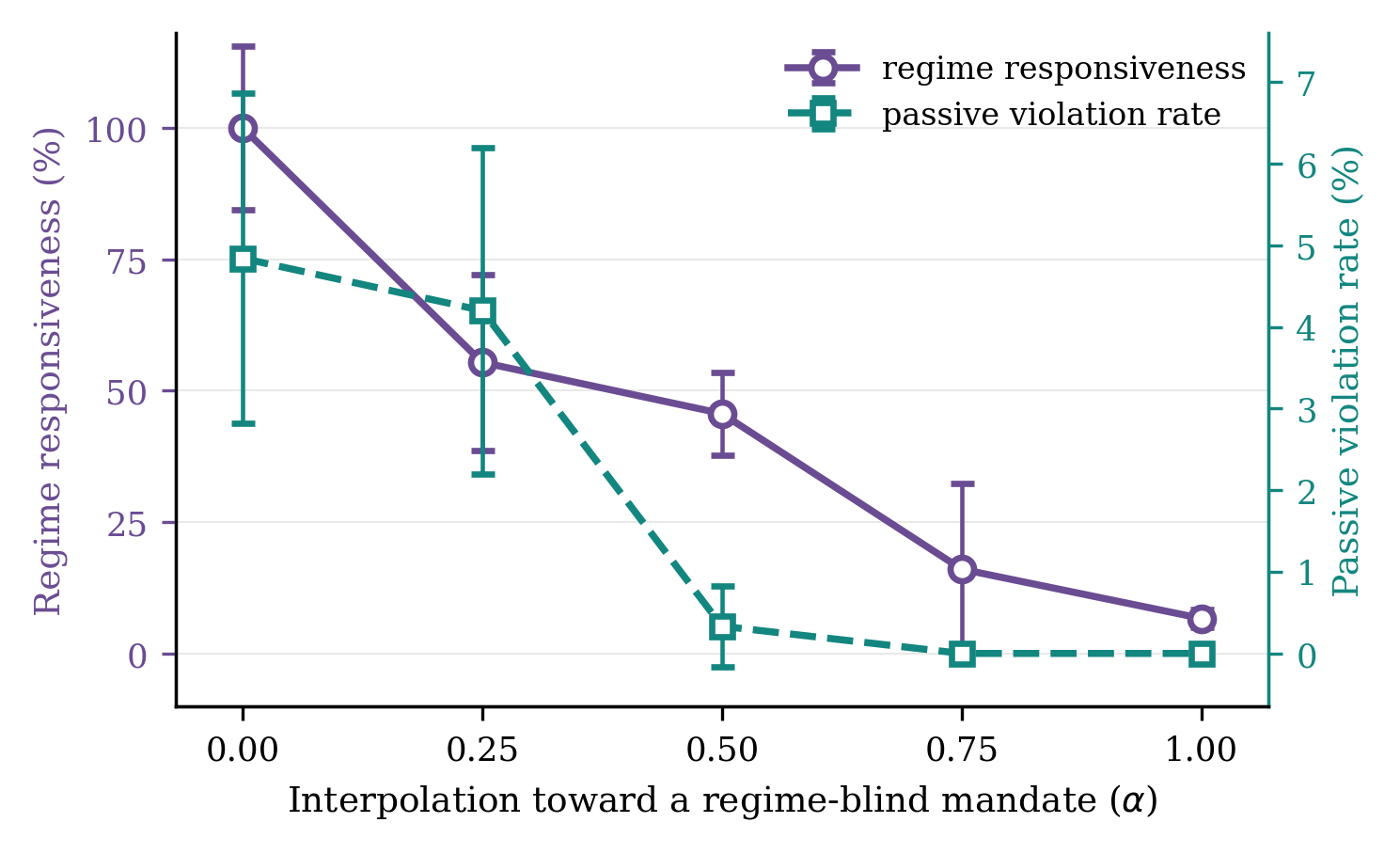}
  \caption{Safe-harbor behavior.}
  \Description{Plot showing regime responsiveness falling and violations dropping to zero as the per-regime safe sets overlap more.}
  \label{fig:harbor}
\end{figure}

\subsection{Safe-Set Overlap and Compliance}
\label{sec:results-harbor}
We interpolate the mandate from the disjoint rule set of Table~\ref{tab:mandate} toward a regime-blind one, over four rule sets of increasing safe-set overlap, with the safety term and its weight $\lambda$ held fixed. Regime responsiveness falls monotonically as the overlap grows (Figure~\ref{fig:harbor}): the policy retains 55\%, 46\%, 16\% and 7\% of its between-regime reallocation across the four sets. The $L_1$ distance behind those percentages falls from 0.409 to 0.028, the spread in equity exposure across regimes from 0.264 to 0.01, and mean temporal dispersion per asset inside a regime from 0.018 to 0.003, so the policy stops moving between regimes and within them together. Compliance moves the other way: passive violations hold at $4.84 \pm 2.03\%$ and $4.20 \pm 2.00\%$ for the first two rule sets, then fall to 0.33\% and to zero on every test day for the last two. Permissiveness rises along the same axis: on $\Delta^{9}$ the mean safe volume per regime goes from 33.6\% of the simplex to 44.0\%, and the Crisis set, where the violation rate is highest, from 6.4\% to 44.0\%.

\section{Discussion and Limitations}
\label{sec:discussion}

This section examines what internalization offers over reward penalties, the safe-harbor effect and its implications for mandate design, then limitations.

\subsection{Internalized Compliance}
A symbolic mandate can be carried out in a continuous-action policy instead of runtime intervention, which is the main result. The policy is substantially compliant without an enforcement layer modifying its deployment actions. The comparison with the reward penalty also illustrates a difference in what the two methods learn. A reward penalty treats compliance as one object among many, so a violation becomes acceptable whenever the extra return covers it, which is what constrained MDP formulations are built to produce. Our test period favored equities while the mandate restricted equity exposure under stress, so remaining overweight often gave the penalty a better objective value than compliance did.The symbolic loss promotes mandate satisfaction directly rather than trading violations against return. Lower returns therefore reflect compliance with a conservative mandate during a market that favored the unconstrained allocation.

In a production environment a runtime shield is non-negotiable because a mandate breach is simply not survivable. However a traditional unconstrained policy essentially "fights" the shield which leads to heavy projection distortions. By internalizing the rules to a 4.8\% baseline error rate, our method acts as a safer base policy, reduces the burden on the runtime enforcement layer and ensures the executed action closely matches the policy's intent.

\subsection{The Safe-Harbor Effect}
We encountered an emergent behavior during preliminary experiments with an earlier version of the mandate: the learned policies barely changed their weights as the regime label changed, producing nearly identical books across regimes. Replacing those thresholds with the disjoint set of Table~\ref{tab:mandate} restored the differentiation with the safety term and its weight unchanged, so what differed was the geometry of the safe sets. Where the per-regime safe sets share a common region, one allocation satisfies several regimes at once, the safety gradient stops depending on which regime is active, and regime-dependent reallocation becomes optional. We call this the \emph{safe-harbor effect}. Related structure is known elsewhere. In policies conditioned on a context that gain little from it when the goals overlap~\cite{beukman2023context}, and in robust optimization, where the value of adapting to the realized scenario turns on whether one decision is feasible across all of them~\cite{bental2004adjustable}. The pull is visible even under the disjoint mandate: in Bull, where the bands are widest, the policy sits essentially on the feasible box's centroid, and in Bear and Crisis, where they are tight, it is pressed against a boundary (the $L_1$ distances of \S\ref{sec:results-harbor}). The policy favors the feasible-set interior and separates regime allocations only when required by their geometry.
 
Redesigning the mandate, as we did, is not a complete repair. A mandate is written by the institution that imposes it and its bands may overlap. Trained against such a mandate the method here yields a policy that complies almost perfectly while conditioning on the regime hardly at all. No compliance metric reveals it, since the parked policy satisfies every rule in every regime. Testing whether the per-regime safe sets intersect costs nothing and can be run before training. A milder form appears in the mandate of Table~\ref{tab:mandate} itself, where equal weight satisfies both the Bull and the Bear bands, so the rule-following heuristic of \S\ref{sec:results-violations} returns the same $50/30/20$ book in both and changes its allocation only on entry to Crisis.

\subsection{Limitations}
\paragraph{The surrogate.} 
The sigmoid at $\beta = 50$ relaxes an indicator and the Monte Carlo logarithm is biased downward by Jensen. We claim a surrogate relationship and not the equivalence of objectives the discrete construction enjoys.
\paragraph{The residual.} 
Compliance stops at $4.8\%$. A third of the Crisis residual is detector latency, the rest sits in the settled crisis book (on 97 test days) with seed spread too wide to order the variants.
\paragraph{Exploration.} 
The entropy coefficient holds $\sigma_\theta \approx 1.5$ and the executed action is noise-dominated, so the return gradient with respect to $\mu_\theta$ is attenuated and the mean never differentiates within a sleeve. Within-sleeve composition is left to the return signal, and on this universe at a daily horizon that signal is weak. The in-sample Sharpe-optimal allocation inside each regime box returns $31.0\%$ at Sharpe $0.51$ out of sample, against $58.1\%$ at $0.80$ for the un-optimized centroid. Compliance is therefore a property of the deployed mean action. Because stochastic sampling increases violations at this entropy, the guarantee covers only the deterministic greedy policy.
\paragraph{Scope.} 
One universe, one mandate, one test window. The environment supports transaction costs and slippage but both are set to zero to separate the agent's ability to internalize the rules from market microstructure and turnover penalties. We study compliance capacity, not a production-ready trading strategy. The unconstrained policies trade far more than the constrained ones, so transaction costs would penalize them more heavily, so their omission only strengthens our comparison. In general, however, the absolute return figures are optimistic throughout.

\section{Related Work}
\label{sec:related}

Deep reinforcement learning for portfolio allocation dates to the introduction of the simplex-valued policy head~\cite{jiang2017drl}, which the standard frameworks in this venue inherit~\cite{liu2021finrl}. Budget and non-negativity come free from the parameterization, since a softmax output cannot represent an infeasible book. Winkel et al.~\cite{winkel2023adbo, winkel2024simplex} extend this to sleeve-level floors and caps, decomposing the constrained polytope into padded simplices so feasibility then belongs to the action space and the agent is never asked to hold a view about the rules at all. Where feasibility cannot be built in, methods for making a reinforcement learner respect a constraint differ mainly in where they put the constraint. Reward shaping subtracts a violation cost from the return~\cite{tessler2019rcpo}, so the constraint reaches the policy through the value function, discounted and averaged over trajectories. Constrained MDP methods make the same trade explicit, bounding an expected discounted cost with a Lagrange multiplier ~\cite{altman1999cmdp} or a trust-region step ~\cite{achiam2017cpo}, and state-wise variants tighten that bound toward per-step guarantees~\cite{zhao2024statewise}. Enkhsaikhan and Jo~\cite{enkhsaikhan2025augmented} carry the augmented-Lagrangian form directly into portfolio allocation, the closest constrained-RL treatment of our problem. In all of them an expectation can hold while any particular day violates. Safety layers ~\cite{dalal2018safe} and shields~\cite{alshiekh2018shielding} correct the action, so the guarantee belongs to the correcting layer and leaves with it.

This is where neurosymbolic learning comes in. A mandate is a known function of the executed weights and neurosymbolic learning exploits that directly. The one neurosymbolic reinforcement learner in this domain feeds symbolic structure to the network as input~\cite{harini2023neurosymbolic}, informing the policy without constraining its actions. Semantic loss derives a differentiable penalty from the probability that a Boolean constraint is satisfied under a network's output distribution~\cite{xu2018semanticloss}. This places the specification directly in the loss, avoiding separate reward or correction layers. PLPG~\cite{yang2023plpg} reweights each action by its probability of safety under a probabilistic logic program and trains through the resulting shielded policy, which needs a normalizer that sums over a finite action set. Both place the specification in a differentiable objective, and both treat the individual rules as independent, which they are not when the rules share variables~\cite{vankrieken2024independence}. Carr et al.~\cite{carr2023shielding} train with an intercepting shield and disable it afterwards, observing persistent safety. We explore this when the shield never enters the action path.

\section{Conclusion}
\label{sec:conclusion}

An investment mandate limits what a portfolio can hold and is typically enforced at the trade, after the policy has acted. We demonstrate that it can be carried within the policy, satisfied by the weights selected by the agent. The construction adapts probabilistic logic shielding to continuous control through a reparameterized Monte Carlo surrogate evaluated on the policy and it requires no world model: for a price-taker, whether an executed allocation satisfies a rule is a closed-form geometric fact. Audited without any enforcement layer, the trained policy violates a six-rule mandate on $4.8\%$ of held out days, against $96$--$98\%$ for unconstrained learners given identical observations over five seeds.

The differentiable safety term accounts for that result. A standard reward penalty, given the same observations and budget, reaches only $33.4\%$. What remains lives at regime boundaries and in the crisis book, roughly a third of it traceable to the detector's latency. The compliant policy earns what a mandated balanced allocation earned over this window, which is the price of a bond floor written for a market that changed. Whether a context-conditioned mandate shapes behavior at all is a question of geometry. Where the per-regime safe sets overlap, one book satisfies several regimes and the conditioning does nothing.

Because the safety term reads only the executed weights, several extensions follow directly. Transaction costs can be incorporated through turnover, while improved regime inference could reduce transition violations caused by detector lag. Extending the mandate to CVaR would test whether internalization can accommodate distributional risk constraints. Allowing mandate thresholds to vary with state would enable the rules themselves to evolve over time, while parameterizing them by risk tolerance could yield a family of policies for different mandate levels. Once a rule can be written down and differentiated, an agent can be made to carry it in its own policy instead of being corrected into it after the fact.


\bibliographystyle{ACM-Reference-Format}
\bibliography{references}


\begin{thebibliography}{32}


\ifx \showCODEN    \undefined \def \showCODEN     #1{\unskip}     \fi
\ifx \showISBNx    \undefined \def \showISBNx     #1{\unskip}     \fi
\ifx \showISBNxiii \undefined \def \showISBNxiii  #1{\unskip}     \fi
\ifx \showISSN     \undefined \def \showISSN      #1{\unskip}     \fi
\ifx \showLCCN     \undefined \def \showLCCN      #1{\unskip}     \fi
\ifx \shownote     \undefined \def \shownote      #1{#1}          \fi
\ifx \showarticletitle \undefined \def \showarticletitle #1{#1}   \fi
\ifx \showURL      \undefined \def \showURL       {\relax}        \fi
\providecommand\bibfield[2]{#2}
\providecommand\bibinfo[2]{#2}
\providecommand\natexlab[1]{#1}
\providecommand\showeprint[2][]{arXiv:#2}

\bibitem[Achiam et~al\mbox{.}(2017)]%
        {achiam2017cpo}
\bibfield{author}{\bibinfo{person}{Joshua Achiam}, \bibinfo{person}{David Held}, \bibinfo{person}{Aviv Tamar}, {and} \bibinfo{person}{Pieter Abbeel}.} \bibinfo{year}{2017}\natexlab{}.
\newblock \showarticletitle{Constrained Policy Optimization}. In \bibinfo{booktitle}{\emph{Proceedings of the 34th International Conference on Machine Learning - Volume 70}} (Sydney, NSW, Australia) \emph{(\bibinfo{series}{ICML'17})}. \bibinfo{publisher}{JMLR.org}, \bibinfo{pages}{22–31}.
\newblock


\bibitem[Alshiekh et~al\mbox{.}(2018)]%
        {alshiekh2018shielding}
\bibfield{author}{\bibinfo{person}{Mohammed Alshiekh}, \bibinfo{person}{Roderick Bloem}, \bibinfo{person}{R{\"u}diger Ehlers}, \bibinfo{person}{Bettina K{\"o}nighofer}, \bibinfo{person}{Scott Niekum}, {and} \bibinfo{person}{Ufuk Topcu}.} \bibinfo{year}{2018}\natexlab{}.
\newblock \showarticletitle{Safe reinforcement learning via shielding}. In \bibinfo{booktitle}{\emph{Proceedings of the Thirty-Second AAAI Conference on Artificial Intelligence}} (New Orleans, Louisiana, USA) \emph{(\bibinfo{series}{AAAI'18})}. \bibinfo{publisher}{AAAI Press}, Article \bibinfo{articleno}{326}, \bibinfo{numpages}{10}~pages.
\newblock
\showISBNx{978-1-57735-800-8}


\bibitem[Altman(1999)]%
        {altman1999cmdp}
\bibfield{author}{\bibinfo{person}{Eitan Altman}.} \bibinfo{year}{1999}\natexlab{}.
\newblock \bibinfo{booktitle}{\emph{Constrained Markov Decision Processes}}.
\newblock \bibinfo{publisher}{CRC Press}.
\newblock


\bibitem[Ang and Bekaert(2002)]%
        {angbekaert2002regime}
\bibfield{author}{\bibinfo{person}{Andrew Ang} {and} \bibinfo{person}{Geert Bekaert}.} \bibinfo{year}{2002}\natexlab{}.
\newblock \showarticletitle{International Asset Allocation With Regime Shifts}.
\newblock \bibinfo{journal}{\emph{The Review of Financial Studies}} \bibinfo{volume}{15}, \bibinfo{number}{4} (\bibinfo{date}{07} \bibinfo{year}{2002}), \bibinfo{pages}{1137--1187}.
\newblock
\showISSN{0893-9454}
\href{https://doi.org/10.1093/rfs/15.4.1137}{doi:\nolinkurl{10.1093/rfs/15.4.1137}}


\bibitem[Ben-Tal et~al\mbox{.}(2004)]%
        {bental2004adjustable}
\bibfield{author}{\bibinfo{person}{Aharon Ben-Tal}, \bibinfo{person}{Alexander Goryashko}, \bibinfo{person}{E. Guslitzer}, {and} \bibinfo{person}{Arkadi Nemirovski}.} \bibinfo{year}{2004}\natexlab{}.
\newblock \showarticletitle{Adjustable robust solutions of uncertain linear programs}.
\newblock \bibinfo{journal}{\emph{Mathematical Programming}}  \bibinfo{volume}{99} (\bibinfo{date}{03} \bibinfo{year}{2004}), \bibinfo{pages}{351--376}.
\newblock
\href{https://doi.org/10.1007/s10107-003-0454-y}{doi:\nolinkurl{10.1007/s10107-003-0454-y}}


\bibitem[Beukman et~al\mbox{.}(2023)]%
        {beukman2023context}
\bibfield{author}{\bibinfo{person}{Michael Beukman}, \bibinfo{person}{Devon Jarvis}, \bibinfo{person}{Richard Klein}, \bibinfo{person}{Steven James}, {and} \bibinfo{person}{Benjamin Rosman}.} \bibinfo{year}{2023}\natexlab{}.
\newblock \showarticletitle{Dynamics Generalisation in Reinforcement Learning via Adaptive Context-Aware Policies}. In \bibinfo{booktitle}{\emph{Thirty-seventh Conference on Neural Information Processing Systems}}.
\newblock
\urldef\tempurl%
\url{https://openreview.net/forum?id=PJhjkSFlbG}
\showURL{%
\tempurl}


\bibitem[Bouneffouf and Aggarwal(2022)]%
        {bouneffouf2022neurosymbolic}
\bibfield{author}{\bibinfo{person}{Djallel Bouneffouf} {and} \bibinfo{person}{Charu~C. Aggarwal}.} \bibinfo{year}{2022}\natexlab{}.
\newblock \bibinfo{title}{Survey on Applications of Neurosymbolic Artificial Intelligence}.
\newblock
\showeprint[arxiv]{2209.12618}~[cs.AI]


\bibitem[Carr et~al\mbox{.}(2023)]%
        {carr2023shielding}
\bibfield{author}{\bibinfo{person}{Steven Carr}, \bibinfo{person}{Nils Jansen}, \bibinfo{person}{Sebastian Junges}, {and} \bibinfo{person}{Ufuk Topcu}.} \bibinfo{year}{2023}\natexlab{}.
\newblock \showarticletitle{Safe reinforcement learning via shielding under partial observability}. In \bibinfo{booktitle}{\emph{Proceedings of the Thirty-Seventh AAAI Conference on Artificial Intelligence}} \emph{(\bibinfo{series}{AAAI'23})}. \bibinfo{publisher}{AAAI Press}, Article \bibinfo{articleno}{1654}, \bibinfo{numpages}{9}~pages.
\newblock
\showISBNx{978-1-57735-880-0}
\href{https://doi.org/10.1609/aaai.v37i12.26723}{doi:\nolinkurl{10.1609/aaai.v37i12.26723}}


\bibitem[Chatterji and Acar(2025)]%
        {chatterji2025analyzing}
\bibfield{author}{\bibinfo{person}{Satchit Chatterji} {and} \bibinfo{person}{Erman Acar}.} \bibinfo{year}{2025}\natexlab{}.
\newblock \showarticletitle{Analyzing Probabilistic Logic Shields for Multi-Agent Reinforcement Learning}.
\newblock In \bibinfo{booktitle}{\emph{ECAI 2025}}. \bibinfo{publisher}{IOS Press}, \bibinfo{pages}{2538--2545}.
\newblock


\bibitem[Dalal et~al\mbox{.}(2018)]%
        {dalal2018safe}
\bibfield{author}{\bibinfo{person}{Gal Dalal}, \bibinfo{person}{Krishnamurthy Dvijotham}, \bibinfo{person}{Matej Vecerik}, \bibinfo{person}{Todd Hester}, \bibinfo{person}{Cosmin Paduraru}, {and} \bibinfo{person}{Yuval Tassa}.} \bibinfo{year}{2018}\natexlab{}.
\newblock \bibinfo{title}{Safe Exploration in Continuous Action Spaces}.
\newblock
\showeprint[arxiv]{1801.08757}~[cs.AI]
\urldef\tempurl%
\url{https://arxiv.org/abs/1801.08757}
\showURL{%
\tempurl}


\bibitem[Enkhsaikhan and Jo(2025)]%
        {enkhsaikhan2025augmented}
\bibfield{author}{\bibinfo{person}{Bayaraa Enkhsaikhan} {and} \bibinfo{person}{Ohyun Jo}.} \bibinfo{year}{2025}\natexlab{}.
\newblock \showarticletitle{Risk-Constrained Reinforcement Learning With Augmented Lagrangian Multiplier for Portfolio Optimization}.
\newblock \bibinfo{journal}{\emph{IEEE Transactions on Big Data}} \bibinfo{volume}{11}, \bibinfo{number}{5} (\bibinfo{year}{2025}), \bibinfo{pages}{2489--2502}.
\newblock


\bibitem[Espiga-Fernández et~al\mbox{.}(2025)]%
        {espiga2025gymfolio}
\bibfield{author}{\bibinfo{person}{Francisco Espiga-Fernández}, \bibinfo{person}{Álvaro García-Sánchez}, {and} \bibinfo{person}{Joaquín Ordieres-Meré}.} \bibinfo{year}{2025}\natexlab{}.
\newblock \showarticletitle{gymfolio: A Reinforcement learning environment for Portfolio Optimization in Python}.
\newblock \bibinfo{journal}{\emph{SoftwareX}}  \bibinfo{volume}{30} (\bibinfo{year}{2025}), \bibinfo{pages}{102106}.
\newblock
\showISSN{2352-7110}
\href{https://doi.org/10.1016/j.softx.2025.102106}{doi:\nolinkurl{10.1016/j.softx.2025.102106}}


\bibitem[{European Parliament and Council}(2009)]%
        {ucits2009}
\bibfield{author}{\bibinfo{person}{{European Parliament and Council}}.} \bibinfo{year}{2009}\natexlab{}.
\newblock \bibinfo{title}{Directive 2009/65/EC on Undertakings for Collective Investment in Transferable Securities ({UCITS})}.
\newblock \bibinfo{howpublished}{Official Journal of the European Union, Art.~52(2)}.
\newblock


\bibitem[Goodall and Belardinelli(2023)]%
        {goodall2023ambs}
\bibfield{author}{\bibinfo{person}{Alexander~W. Goodall} {and} \bibinfo{person}{Francesco Belardinelli}.} \bibinfo{year}{2023}\natexlab{}.
\newblock \bibinfo{booktitle}{\emph{Approximate Model-Based Shielding for Safe Reinforcement Learning}}.
\newblock \bibinfo{publisher}{IOS Press}.
\newblock
\showISBNx{9781643684376}
\showISSN{1879-8314}
\href{https://doi.org/10.3233/faia230357}{doi:\nolinkurl{10.3233/faia230357}}


\bibitem[Goodall and Belardinelli(2024)]%
        {goodall2024continuous}
\bibfield{author}{\bibinfo{person}{Alexander~W. Goodall} {and} \bibinfo{person}{Francesco Belardinelli}.} \bibinfo{year}{2024}\natexlab{}.
\newblock \showarticletitle{Leveraging Approximate Model-based Shielding for Probabilistic Safety Guarantees in Continuous Environments}. In \bibinfo{booktitle}{\emph{Proceedings of the 23rd International Conference on Autonomous Agents and Multiagent Systems (AAMAS)}}. \bibinfo{address}{Richland, SC}, \bibinfo{pages}{2291–2293}.
\newblock
\showISBNx{9798400704864}


\bibitem[Hamilton(1989)]%
        {hamilton1989regime}
\bibfield{author}{\bibinfo{person}{James~D. Hamilton}.} \bibinfo{year}{1989}\natexlab{}.
\newblock \showarticletitle{A New Approach to the Economic Analysis of Nonstationary Time Series and the Business Cycle}.
\newblock \bibinfo{journal}{\emph{Econometrica}} \bibinfo{volume}{57}, \bibinfo{number}{2} (\bibinfo{year}{1989}), \bibinfo{pages}{357--384}.
\newblock
\showISSN{00129682, 14680262}
\urldef\tempurl%
\url{http://www.jstor.org/stable/1912559}
\showURL{%
\tempurl}


\bibitem[Harini et~al\mbox{.}(2023)]%
        {harini2023neurosymbolic}
\bibfield{author}{\bibinfo{person}{S~I Harini}, \bibinfo{person}{Gautam Shroff}, \bibinfo{person}{Ashwin Srinivasan}, \bibinfo{person}{Prayushi Faldu}, {and} \bibinfo{person}{Lovekesh Vig}.} \bibinfo{year}{2023}\natexlab{}.
\newblock \bibinfo{title}{Neuro-symbolic Meta Reinforcement Learning for Trading}.
\newblock
\showeprint[arxiv]{2302.08996}~[cs.AI]
\urldef\tempurl%
\url{https://arxiv.org/abs/2302.08996}
\showURL{%
\tempurl}


\bibitem[Jiang et~al\mbox{.}(2017)]%
        {jiang2017drl}
\bibfield{author}{\bibinfo{person}{Zhengyao Jiang}, \bibinfo{person}{Dixing Xu}, {and} \bibinfo{person}{Jinjun Liang}.} \bibinfo{year}{2017}\natexlab{}.
\newblock \bibinfo{title}{A Deep Reinforcement Learning Framework for the Financial Portfolio Management Problem}.
\newblock
\showeprint[arxiv]{1706.10059}~[q-fin.CP]
\urldef\tempurl%
\url{https://arxiv.org/abs/1706.10059}
\showURL{%
\tempurl}


\bibitem[Kautz(2022)]%
        {kautz2022thirdsummer}
\bibfield{author}{\bibinfo{person}{Henry~A. Kautz}.} \bibinfo{year}{2022}\natexlab{}.
\newblock \showarticletitle{The Third {AI} Summer: {AAAI} Robert S. Engelmore Memorial Lecture}.
\newblock \bibinfo{journal}{\emph{AI Magazine}} \bibinfo{volume}{43}, \bibinfo{number}{1} (\bibinfo{year}{2022}), \bibinfo{pages}{105--125}.
\newblock


\bibitem[Kingma and Welling(2014)]%
        {kingma2014auto}
\bibfield{author}{\bibinfo{person}{Diederik~P Kingma} {and} \bibinfo{person}{Max Welling}.} \bibinfo{year}{2014}\natexlab{}.
\newblock \showarticletitle{Auto-Encoding Variational Bayes}. In \bibinfo{booktitle}{\emph{2nd International Conference on Learning Representations, {ICLR} 2014}}.
\newblock


\bibitem[Liu et~al\mbox{.}(2022)]%
        {liu2021finrl}
\bibfield{author}{\bibinfo{person}{Xiao-Yang Liu}, \bibinfo{person}{Hongyang Yang}, \bibinfo{person}{Jiechao Gao}, {and} \bibinfo{person}{Christina~Dan Wang}.} \bibinfo{year}{2022}\natexlab{}.
\newblock \showarticletitle{FinRL: deep reinforcement learning framework to automate trading in quantitative finance}. In \bibinfo{booktitle}{\emph{Proceedings of the Second ACM International Conference on AI in Finance}} (Virtual Event) \emph{(\bibinfo{series}{ICAIF '21})}. Article \bibinfo{articleno}{1}, \bibinfo{numpages}{9}~pages.
\newblock
\showISBNx{9781450391481}


\bibitem[Maginn et~al\mbox{.}(2007)]%
        {maginn2007managing}
\bibfield{editor}{\bibinfo{person}{John~L. Maginn}, \bibinfo{person}{Donald~L. Tuttle}, \bibinfo{person}{Jerald~E. Pinto}, {and} \bibinfo{person}{Dennis~W. McLeavey}} (Eds.). \bibinfo{year}{2007}\natexlab{}.
\newblock \bibinfo{booktitle}{\emph{Managing Investment Portfolios: A Dynamic Process} (\bibinfo{edition}{3rd} ed.)}.
\newblock


\bibitem[{Office of Financial Research}(2026)]%
        {ofr2026fsi}
\bibfield{author}{\bibinfo{person}{{Office of Financial Research}}.} \bibinfo{year}{2026}\natexlab{}.
\newblock \bibinfo{title}{{OFR} Financial Stress Index}.
\newblock \bibinfo{howpublished}{U.S. Department of the Treasury, refreshed daily}.
\newblock
\shownote{\url{https://www.financialresearch.gov/financial-stress-index/}, accessed 28 July 2026}.
\newblock


\bibitem[Schulman et~al\mbox{.}(2017)]%
        {schulman2017ppo}
\bibfield{author}{\bibinfo{person}{John Schulman}, \bibinfo{person}{Filip Wolski}, \bibinfo{person}{Prafulla Dhariwal}, \bibinfo{person}{Alec Radford}, {and} \bibinfo{person}{Oleg Klimov}.} \bibinfo{year}{2017}\natexlab{}.
\newblock \bibinfo{title}{Proximal Policy Optimization Algorithms}.
\newblock
\showeprint[arxiv]{1707.06347}~[cs.LG]
\urldef\tempurl%
\url{https://arxiv.org/abs/1707.06347}
\showURL{%
\tempurl}


\bibitem[Sutton and Barto(2018)]%
        {sutton2018rl}
\bibfield{author}{\bibinfo{person}{Richard~S. Sutton} {and} \bibinfo{person}{Andrew~G. Barto}.} \bibinfo{year}{2018}\natexlab{}.
\newblock \bibinfo{booktitle}{\emph{Reinforcement Learning: An Introduction} (\bibinfo{edition}{2nd} ed.)}.
\newblock \bibinfo{publisher}{MIT Press}, \bibinfo{address}{Cambridge, MA}.
\newblock


\bibitem[Tessler et~al\mbox{.}(2019)]%
        {tessler2019rcpo}
\bibfield{author}{\bibinfo{person}{Chen Tessler}, \bibinfo{person}{Daniel~J. Mankowitz}, {and} \bibinfo{person}{Shie Mannor}.} \bibinfo{year}{2019}\natexlab{}.
\newblock \showarticletitle{Reward Constrained Policy Optimization}. In \bibinfo{booktitle}{\emph{7th International Conference on Learning Representations, {ICLR} 2019}}.
\newblock
\urldef\tempurl%
\url{https://openreview.net/forum?id=SkfrvsA9FX}
\showURL{%
\tempurl}


\bibitem[Van~Krieken et~al\mbox{.}(2024)]%
        {vankrieken2024independence}
\bibfield{author}{\bibinfo{person}{Emile Van~Krieken}, \bibinfo{person}{Pasquale Minervini}, \bibinfo{person}{Edoardo~M. Ponti}, {and} \bibinfo{person}{Antonio Vergari}.} \bibinfo{year}{2024}\natexlab{}.
\newblock \showarticletitle{On the independence assumption in neurosymbolic learning}. In \bibinfo{booktitle}{\emph{Proceedings of the 41st International Conference on Machine Learning (ICML)}} (Vienna, Austria). \bibinfo{publisher}{JMLR.org}, Article \bibinfo{articleno}{2007}, \bibinfo{numpages}{20}~pages.
\newblock


\bibitem[Winkel et~al\mbox{.}(2023a)]%
        {winkel2023adbo}
\bibfield{author}{\bibinfo{person}{David Winkel}, \bibinfo{person}{Niklas Strau{\ss}}, \bibinfo{person}{Matthias Schubert}, \bibinfo{person}{Yunpu Ma}, {and} \bibinfo{person}{Thomas Seidl}.} \bibinfo{year}{2023}\natexlab{a}.
\newblock \showarticletitle{Constrained Portfolio Management Using Action Space Decomposition for Reinforcement Learning}. In \bibinfo{booktitle}{\emph{Advances in Knowledge Discovery and Data Mining: 27th Pacific-Asia Conference on Knowledge Discovery and Data Mining, PAKDD 2023}}. \bibinfo{publisher}{Springer-Verlag}, \bibinfo{address}{Berlin, Heidelberg}, \bibinfo{pages}{373–385}.
\newblock
\showISBNx{978-3-031-33376-7}


\bibitem[Winkel et~al\mbox{.}(2023b)]%
        {winkel2024simplex}
\bibfield{author}{\bibinfo{person}{David Winkel}, \bibinfo{person}{Niklas Strauß}, \bibinfo{person}{Matthias Schubert}, {and} \bibinfo{person}{Thomas Seidl}.} \bibinfo{year}{2023}\natexlab{b}.
\newblock \bibinfo{booktitle}{\emph{Simplex Decomposition for Portfolio Allocation Constraints in Reinforcement Learning}}.
\newblock \bibinfo{publisher}{IOS Press}.
\newblock
\showISBNx{9781643684376}
\showISSN{1879-8314}


\bibitem[Xu et~al\mbox{.}(2018)]%
        {xu2018semanticloss}
\bibfield{author}{\bibinfo{person}{Jingyi Xu}, \bibinfo{person}{Zilu Zhang}, \bibinfo{person}{Tal Friedman}, \bibinfo{person}{Yitao Liang}, {and} \bibinfo{person}{Guy Van~den Broeck}.} \bibinfo{year}{2018}\natexlab{}.
\newblock \showarticletitle{A Semantic Loss Function for Deep Learning with Symbolic Knowledge}. In \bibinfo{booktitle}{\emph{Proceedings of the 35th International Conference on Machine Learning}} \emph{(\bibinfo{series}{Proceedings of Machine Learning Research}, Vol.~\bibinfo{volume}{80})}, \bibfield{editor}{\bibinfo{person}{Jennifer Dy} {and} \bibinfo{person}{Andreas Krause}} (Eds.). \bibinfo{publisher}{PMLR}, \bibinfo{pages}{5502--5511}.
\newblock
\urldef\tempurl%
\url{https://proceedings.mlr.press/v80/xu18h.html}
\showURL{%
\tempurl}


\bibitem[Yang et~al\mbox{.}(2023)]%
        {yang2023plpg}
\bibfield{author}{\bibinfo{person}{Wen-Chi Yang}, \bibinfo{person}{Giuseppe Marra}, \bibinfo{person}{Gavin Rens}, {and} \bibinfo{person}{Luc De~Raedt}.} \bibinfo{year}{2023}\natexlab{}.
\newblock \showarticletitle{Safe Reinforcement Learning via Probabilistic Logic Shields}. In \bibinfo{booktitle}{\emph{Proceedings of the 32nd International Joint Conference on Artificial Intelligence (IJCAI-23)}}. \bibinfo{pages}{5739--5749}.
\newblock


\bibitem[Zhao et~al\mbox{.}(2024)]%
        {zhao2024statewise}
\bibfield{author}{\bibinfo{person}{Weiye Zhao}, \bibinfo{person}{Rui Chen}, \bibinfo{person}{Yifan Sun}, \bibinfo{person}{Tianhao Wei}, {and} \bibinfo{person}{Changliu Liu}.} \bibinfo{year}{2024}\natexlab{}.
\newblock \bibinfo{title}{State-wise Constrained Policy Optimization}.
\newblock
\showeprint[arxiv]{2306.12594}~[cs.LG]
\urldef\tempurl%
\url{https://arxiv.org/abs/2306.12594}
\showURL{%
\tempurl}


\end{thebibliography}

\end{document}